%% file: neurips_2026_vericode_workshop.tex
\documentclass{article}

\usepackage[main, final]{neurips_2026_vericode}

\usepackage[utf8]{inputenc} % allow utf-8 input
\usepackage[T1]{fontenc}    % use 8-bit T1 fonts
\usepackage{hyperref}       % hyperlinks
\usepackage{url}            % simple URL typesetting
\usepackage{booktabs}       % professional-quality tables
\usepackage{amsfonts}       % blackboard math symbols
\usepackage{nicefrac}       % compact symbols for 1/2, etc.
\usepackage{microtype}      % microtypography
\usepackage{xcolor}         % colors

\usepackage{graphicx}
\usepackage{subcaption}
\usepackage{enumitem}
\usepackage{multirow}
\usepackage{wrapfig}
\usepackage{minted}
\usepackage{listings}

\usepackage[most]{tcolorbox}

\lstdefinestyle{smt}{
  basicstyle=\ttfamily\footnotesize,
  breaklines=true,
  frame=none,
  language=Lisp
}

\usepackage{algorithm}
\usepackage{algpseudocode}

\definecolor{codegray}{gray}{0.95}
\usepackage{amsmath}
\usepackage{amssymb}
\usepackage{mathtools}
\usepackage{amsthm}
\usepackage{bm}
\usepackage[capitalize,noabbrev]{cleveref}
\usepackage{lipsum}

\theoremstyle{plain}

\theoremstyle{definition}

\theoremstyle{remark}

\usepackage[textsize=tiny]{todonotes}
\usepackage{caption}
\usepackage{wrapfig}

\input{macros.tex}

\newcommand{\eat}[1]{}

\title{
  Stratified Consistency Distillation for\\ Natural Language Formalization
}

\author{
  Zhichao Hou\textsuperscript{\rm 1}
  \And
  Ferhat Erata\textsuperscript{\rm 2} 
  \And
  Joe Lilien\textsuperscript{\rm 2}
  \And
  MohamadAli Torkamani\textsuperscript{\rm 2}
  \\
    \and
  \textsuperscript{\rm 1}North Carolina State University, 
  \textsuperscript{\rm 2}Amazon Web Services\\
  \and
  \texttt{zhou4@ncsu.edu, 
  \{erata,lilienj,alitor\}@amazon.com}\\
}

\begin{document}

\maketitle

\input{sections/abs}
\input{sections/intro}

\input{sections/pre}
\input{sections/method}

\input{sections/exp}
\input{sections/related}

\input{sections/conclusion}

\newpage
% \section*{References}

% \nocite{langley00}

\bibliography{example_paper}
\bibliographystyle{icml2024}

%%%%%%%%%%%%%%%%%%%%%%%%%%%%%%%%%%%%%%%%%%%%%%%%%%%%%%%%%%%%

\appendix
\input{sections/app}
%%%%%%%%%%%%%%%%%%%%%%%%%%%%%%%%%%%%%%%%%%%%%%%%%%%%%%%%%%%%

% \newpage
% \input{checklist.tex}

\end{document}

%% file: macros.tex
\usepackage{mathtools}
\usepackage{amssymb}
\usepackage{amsthm}
\usepackage{color}
\newcommand{\cut}[1]{{}} 
\newcommand{\vp}{{\mathbf{p}}}

\newcommand{\vt}{{\mathbf{t}}}

\makeatletter
\let\@@span\span
\def\sp@n{\@@span\omit\advance\@multicnt\m@ne}
\makeatother

\newcommand{\bc}{\begin{center}}
\newcommand{\ec}{\end{center}}

\newcommand{\bdm}{\begin{displaymath}}
\newcommand{\edm}{\end{displaymath}}

\newcommand{\beq}{\begin{equation}}
\newcommand{\eeq}{\end{equation}}

\newcommand{\bfl}{\begin{flushleft}}
\newcommand{\efl}{\end{flushleft}}

\newcommand{\bt}{\begin{tabbing}}
\newcommand{\et}{\end{tabbing}}

\newcommand{\beqn}{\begin{align}}
\newcommand{\eeqn}{\end{align}}

\newcommand{\beqs}{\begin{align*}} % no equation numbers
\newcommand{\eeqs}{\end{align*}}  % no equation numbers

%% file: sections/abs.tex
\begin{abstract}
Neurosymbolic reasoning has shown promising success in addressing complex reasoning tasks by combining large language models (LLMs) and symbolic solvers. While this approach shows promise, a fundamental challenge remains: improving the accuracy of translations from natural language to logical formulas. Current methods predominantly rely on prompt engineering, which is difficult to scale across different domains and input formats. Drawing inspiration from the success of fine-tuning in other model adaptation and alignment applications, we propose a fine-tuning-based Stratified Consistency Distillation approach: (1) We generate K logical translations per input using a frontier LLM and cluster them by semantic equivalence (2) Based on the entropy level, we apply majority voting (low entropy), LLM-as-a-Judge (medium entropy), or unification/abstention (high entropy), and (3) fine-tune a smaller model using the selected pseudo-labels.  Our  experiments show significant and consistent improvements in both Pass@K and our novel Equivalent Logical Similarity metrics, demonstrating the potential of advancing logical translation through consistency distillation.
\eat{Neurosymbolic reasoning has shown promising success in addressing complex reasoning tasks by combining the strengths of large language models (LLMs) and symbolic solvers. A core challenge, however, is how to efficiently and effectively improve the translation accuracy from natural language to logical formulas. Existing methods rely heavily on prompt engineering, which does not scale well across diverse domains and input formats. Motivated by the benefits of fine-tuning in various applications, we propose a more systematic fine-tuning–based approach that can be incrementally integrated into existing systems through two pipelines: (1) Stratified Consistency Distillation: We generate 10 SMT-LIB translations per input using a frontier LLM and cluster them by semantic equivalence. Based on entropy levels, we apply majority voting (low), LLM-as-a-Judge (mid), or unification/abstention (high), and fine-tune a smaller model using the selected pseudo-labels. (2) Equivalent-Augmented Fine-Tuning: We improve SMT-LIB implication translation by first adding direct SMT-LIB data for auxiliary learning. Then, we synthesize and rank equivalent examples by AR probability, and use the least confident ones for continued learning. Our comprehensive experiments demonstrate significant and consistent improvements in both Pass@K and our novel Equivalent Similarity metrics, highlighting a promising direction for advancing logical translation through consistency distillation, equivalent augmentation, and fine-tuning.}
\end{abstract}

%% file: sections/intro.tex
\section{Introduction}
\label{sec:intro}

Large Language Models (LLMs) have become a cornerstone technology for building customer-facing chatbot systems, enabling natural, context-aware, and highly interactive conversations across diverse application domains~\citep{hou2026tefm,zhang2026centile,wang2025exploratory}. 
Despite their impressive capabilities, LLMs are inherently prone to hallucinations—the generation of factually incorrect or logically inconsistent outputs—which may directly contradict authoritative source-of-truth documents. 
Such errors can be catastrophic in high-stakes domains such as compliance, pricing, and regulations, where a single incorrect response may result in financial loss, legal liability, or reputational damage~\citep{hou2024protransformer,hou2025robustness,hou2026towards}.

A principled way to ensure the \emph{logical soundness} and \emph{policy compliance} of chatbot responses is to convert human-written natural language into \emph{verifiable formal representations} (e.g., SMT-LIB) and to verify their correctness using automated reasoning tools such as the Z3 solver. 
This leads us to a fundamental research question in the emerging field of \emph{LLM reasoning and autoformalization}:

\begin{quote}
    \textit{Can LLMs accurately understand human-written natural language and translate it into formal logical formulas that can be verified by a symbolic solver?}
\end{quote}

Recent advances in frontier LLMs~\citep{openai2024gpt4technicalreport,anthropic2024claude3,geminiteam2024gemini15unlockingmultimodal}—empowered by few-shot in-context learning and chain-of-thought prompting~\citep{fu2023complexitybasedpromptingmultistepreasoning,wei2023chainofthoughtpromptingelicitsreasoning}—have demonstrated remarkable capabilities in such translation tasks. 
However, prompt-based approaches using frontier LLMs face two major limitations: 
(1) Frontier models are extremely large, with parameter counts ranging from $70$B to over $500$B, leading to high inference latency and prohibitive computational cost~\citep{gao2025modulated}; and 
(2) These models are generally closed-source and only available through black-box APIs, preventing fine-tuning and constraining performance improvements beyond prompt engineering.

To address these limitations, we propose a systematic and scalable framework—\emph{Stratified Consistency Distillation}—for translating policy-governed natural language into formal logic, which distills the reasoning and translation capabilities of a frontier LLM into a smaller, more efficient open-source language model.
Our main contributions are summarized as follows:

\begin{itemize}[left=0.0em]
    \item \textbf{Synthetic Dataset Generation from Policy Documents.}  
    We design a robust pipeline to automatically extract domain-relevant rules from unstructured policy documents and generate aligned natural language–SMT-LIB training pairs, enabling scalable data creation without manual annotation.
    
    \item \textbf{Stratified Consistency Distillation (SCD).}  
    We introduce a novel distillation strategy that leverages semantic equivalence clustering and entropy-based stratification to selectively transfer knowledge from a high-capability frontier LLM into a smaller LM, preserving logical consistency while reducing inference cost.
        
    \item \textbf{Comprehensive Evaluation and Analysis.}  
    We conduct extensive experiments across multiple policy-driven reasoning benchmarks. Our method achieves substantial improvements in logical translation accuracy over both prompt-based frontier LLM approaches and fine-tuned baselines, while offering $5\times$–$20\times$ lower inference cost.
    
    % \item \textbf{Practical Implications.}  
    % Beyond the academic contribution, our framework provides a practical blueprint for deploying policy-compliant chatbot systems in regulated industries, demonstrating that high reasoning performance can be retained without reliance on proprietary large-scale models.
\end{itemize}

%% file: sections/pre.tex
% ```latex
\section{Preliminary}

As large language models (LLMs) are increasingly used to power customer-facing chatbots, ensuring that their outputs remain accurate and aligned with formal company policies has become a critical challenge, especially in high-stakes domains such as compliance, pricing, and regulation. Despite their impressive capabilities, LLMs may generate hallucinated or logically inconsistent responses that contradict authoritative source documents. To improve chatbot reliability, we combine the natural language understanding capabilities of LLMs with formal logical reasoning, ensuring that generated responses are not only fluent but also logically sound and policy-compliant. In this section, we introduce our pipeline for generating synthetic data from policy documents, the resulting SMT-LIB representation, the NL2SMT problem formulation, and the evaluation metrics.

\begin{figure}[h!]
    \centering
    \vspace{-0.1in}
    \includegraphics[width=1.0\linewidth]{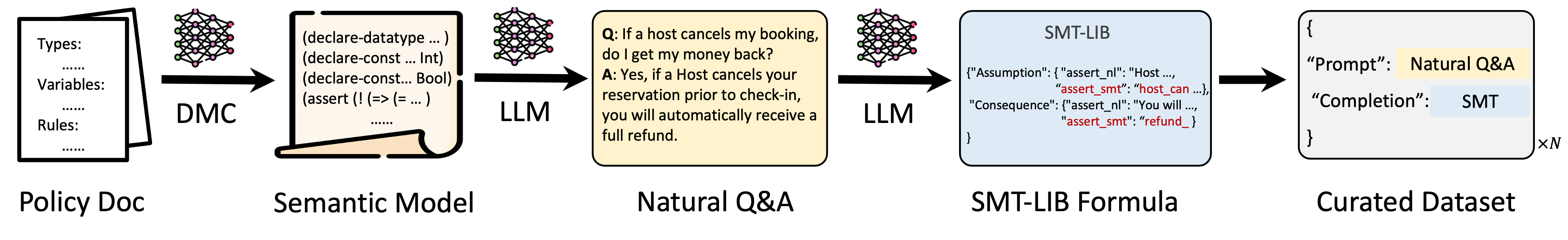}
    \caption{Synthetic dataset generation pipeline.}
    \vspace{-0.1in}
    \label{fig:data_generation}
\end{figure}

\textbf{Synthetic dataset generation from policy documents.}
We generate natural language question-answer pairs and their corresponding SMT-LIB translations using information extracted from source policy documents. The pipeline consists of the following steps, as illustrated in Figure~\ref{fig:data_generation}:

\begin{enumerate}[left=0.0em]
    \item \textit{Extract semantic context.}
    The process begins with a source policy document. An LLM is used to construct a semantic model of the document, including an SMT-LIB specification consisting of relevant declarations, variables, and policy rules. The extracted declarations and contextual information guide the subsequent data-generation steps.

    \item \textit{Generate natural language Q\&A pairs.}
    An LLM generates natural language question-answer pairs based on the extracted semantic context. These pairs represent realistic user-chatbot interactions and are grounded in the content and rules of the source policy document.

    \item \textit{Translate natural language into SMT-LIB.}
    The generated natural language content and the extracted semantic context are incorporated into the translation prompt. Using its reasoning capabilities, the LLM translates each natural language instance into a complete and syntactically valid SMT-LIB representation that captures the logical semantics of the question-answer pair.

    \item \textit{Construct the final training pairs.}
    Each generated example is organized in the following format:
    \[
    \texttt{\textless Natural Language Prompt\textgreater{}}
    \;\rightarrow{}\;
    \texttt{\textless SMT-LIB Completion\textgreater{}}.
    \]
    These prompt-completion pairs form the training dataset used to align natural language inputs with their formal logical representations.
\end{enumerate}

\textbf{SMT-LIB data representation.}
The resulting dataset consists of natural language prompts paired with their corresponding SMT-LIB translations. Each translation expresses the logical content of the input using the declarations, variables, and rules extracted from the relevant semantic context. Depending on the input, an SMT-LIB completion may contain logical operators, quantified expressions, constraints, or relationships between conditions and their consequences. All examples nevertheless follow a unified SMT-LIB representation and are treated as instances of the same NL2SMT translation task.

\textbf{NL2SMT problem setup.}
We aim to improve the translation of natural language inputs into formal SMT-LIB representations. Given a natural language prompt $\vp$, which may include task instructions, contextual information, and a question-answer pair, the goal is to generate an SMT-LIB completion $\vt$ that accurately captures its logical semantics.
Formally, we define a training dataset $\mathcal{D}$ consisting of paired samples $(\vp,\vt)$, where $\vt$ is the ground-truth SMT-LIB translation of $\vp$. Let $\mathbb{LM}_{\theta}$ denote a parameterized language model with parameters $\theta$. Our objective is to maximize the expected similarity $\mathcal{S}$ between the generated translation $\mathbb{LM}_{\theta}(\vp)$ and the ground-truth translation $\vt$, where $\mathcal{S}$ measures either exact logical equivalence or continuous logical similarity:
\begin{equation}
% \vspace{-0.01in}
\max_{\theta}
\mathbb{E}_{(\vp,\vt)\sim\mathcal{D}}
\left[
\mathcal{S}\left(\mathbb{LM}_{\theta}(\vp),\vt\right)
\right].
\label{eq:problem}
\end{equation}
\textbf{Equivalence measurements.}
To evaluate the quality of the generated SMT-LIB translations, we consider two complementary measurements of $\mathcal{S}$:

\begin{itemize}[left=0.0em]
    \item \textit{Binary Equivalence Check.}
    We use the Z3 theorem prover to determine whether a generated SMT-LIB formula is logically equivalent to its ground-truth counterpart. The check returns \texttt{True} when the two formulas are equivalent and \texttt{False} otherwise. We report \textbf{Pass@10}, which measures whether at least one of ten generated candidates is logically equivalent to the ground-truth formula.

    \item \textit{Continuous Similarity Score.}
    When exact equivalence is not achieved, we compute a graded similarity score in the range $[0,1]$ using structural compression and anti-unification over the generated and ground-truth SMT-LIB formulas. Specifically, we use Egglog to extract an anti-unifier according to the defined SMT-LIB specification. The resulting score measures the degree of shared logical structure between the formulas.
\end{itemize}

Together, these measurements provide both a strict assessment of logical equivalence and a continuous assessment of partial logical alignment between generated and ground-truth SMT-LIB formulas.
% ```

%% file: sections/method.tex
% \newpage
\section{Stratified Consistency Distillation}
\label{sec:method}
\begin{figure}[h!]
    \centering
    \includegraphics[width=1.0\linewidth]{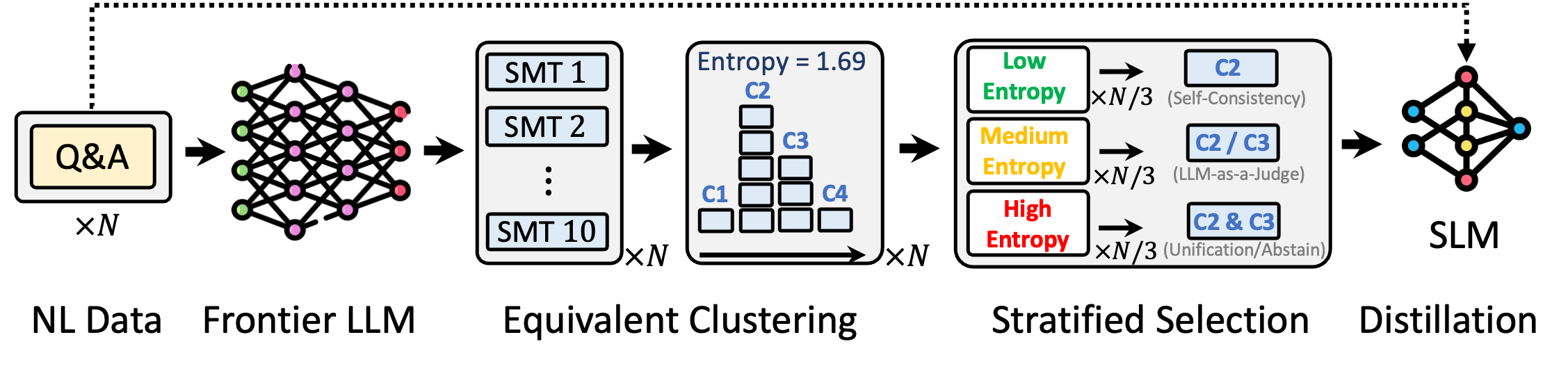}
    \caption{
    Schematic overview of Stratified Consistency Distillation: For each training sample, we generate 10 SMT-LIB translations using a frontier LLM (e.g., Claude Sonnet 3.7). These translations are clustered based on semantic equivalence, and semantic entropy is computed. The data is then stratified into three groups by entropy: 
    \textit{Low entropy}: Apply majority voting for self-consistency; 
    \textit{Medium entropy}: Use LLM-as-a-Judge to select among the top-2 clusters; 
    \textit{High entropy}: Perform unification of the top-2 translations or abstain the data.
    Finally, we distill the knowledge into a smaller model (e.g., Qwen) via fine-tuning on Q\&A pairs and pseudo-labels derived from stratified selection. 
    }
    \label{fig:overview}
    \vspace{-0.2in}
\end{figure}

In this section, we introduce a systematic framework for logical translation, Stratified Consistency Distillation, which transfers knowledge from a frontier LLM to a smaller LM. The systematic overview of our framework is provided in Figure~\ref{fig:overview}.

\subsection{Stratified Consistency Distillation} 

The pretrained frontier large language models (LLMs)~\citep{openai2024gpt4technicalreport,anthropic2024claude3,geminiteam2024gemini15unlockingmultimodal} have demonstrated superior reasoning capabilities, especially when combined with prompting techniques such as chain-of-thought~\citep{fu2023complexitybasedpromptingmultistepreasoning, wei2023chainofthoughtpromptingelicitsreasoning}. However, deploying these powerful LLMs in real-world scenarios remains challenging due to two main limitations: (1) They are typically extremely large, with model sizes ranging from 70B to over 500B parameters, resulting in significant latency and computational overhead during inference; and (2) These models are generally closed-source and only accessible through black-box APIs, which prevents fine-tuning and thus limits performance improvements beyond prompt engineering.

To overcome these limitations, we propose a Stratified Consistency Distillation pipeline that transfers knowledge from frontier teacher LLMs to a smaller student generator, as illustrated in Figure~\ref{fig:overview}. At a high level, our pipeline consists of four main steps:
\vspace{-0.1in}
\begin{enumerate}[left=0.0em]
    \item \textbf{Generation:} Sample output sequences of tokens from the predictive distribution of a frontier LLM given an input prompt $\vp$.
    
    \item \textbf{Clustering:} Cluster the generated translations based on their logical equivalence using our proposed clustering algorithm. Estimate semantic entropy by summing the probabilities of clusters, following a defined entropy formulation.
    
    \item \textbf{Selection:} Select pseudo labels using different strategies depending on the estimated entropy.
    
    \item \textbf{Distillation:} Use the input and selected pseudo label from the frontier LLM to train a smaller student model.
\end{enumerate}
\vspace{-0.1in}

\textbf{Redundant Generation.}
Given a prompt $\vp$ (including necessary instructions, few-shot examples, and the Q\&A to be translated), we sample $M$ SMT-LIB translations $\{\vt^{(1)}, \vt^{(2)}, \ldots, \vt^{(M)}\}$ from a frontier LLM (e.g., Claude Sonnet 3.7~\citep{anthropic2024claude3}), denoted as $\mathbb{LLM}$. 
% To accelerate the sampling process, we employ \texttt{vLLM}~\citep{kwon2023efficient}, a high-throughput and memory-efficient inference and serving engine.

\textbf{Equivalent Clustering \& Symbolic Semantic Entropy.}
We invoke our \texttt{check\_smt\_equivalence} function $\mathbb{E}(\cdot,\cdot)$ to group the $M$ translations into equivalence clusters based on logical equivalence.
The \texttt{check\_smt\_equivalence} function verifies whether two SMT-LIB expressions (\texttt{smt1} and \texttt{smt2}) are logically equivalent under given \texttt{declarations} using the Z3 SMT solver. The function constructs implication trees, writes them to an SMT-LIB file, and checks the satisfiability of the negated equality condition. A return value of \texttt{unsat} from Z3 indicates logical equivalence. The function outputs a dictionary containing the equivalence result, 
% (\texttt{True} or \texttt{False}), 
Z3 output, and any errors.
Semantic entropy~\citep{farquhar2024detecting} has been introduced in natural language reasoning as a means to improve the correctness and soundness of LLM-generated reasoning. Extending this concept, we introduce symbolic semantic equivalence as an alternative formulation, described as follows.
Recall that an equivalence relation is reflexive, symmetric, and transitive. Any such relation induces a set of equivalence classes. Each semantic equivalence class groups outputs expressing the same logical meaning. That is, for the set of classes $\mathcal{C}$, all sentences $\vt, \vt' \in \mathbf{c} \in \mathcal{C}$ satisfy $\mathbb{E}(\vt, \vt') = \texttt{True}$. New sentences are added to an existing class if they match any existing member, otherwise a new class is formed. 
Given the clusters $\mathcal{C}$, we compute semantic entropy as:
$\text{SE}(\vp) = -\sum_{i=1}^{|\mathcal{C}|} P(\mathcal{C}_i|\vp) \log P(\mathcal{C}_i|\vp),$
where $P(\mathcal{C}_i|\vp)$ is the normalized probability of cluster $\mathcal{C}_i$.

\textbf{Stratified Selection for Pseudo Labels.}
After computing entropy for each input, we construct a training dataset $\left\{\vp_i, \{\vt_i^{(j)}\}_{j=1}^{M}, e_i\right\}_{i=1}^{N}$, where $e_i$ is the entropy value. We stratify the data into three groups based on entropy and apply different strategies to select pseudo labels:
\begin{itemize}[left=0.0em]
    \item \textit{Low entropy}: The generation distribution is concentrated on a dominant cluster. We select a translation from the largest cluster: $\vt \in \arg\max_{\mathcal{C}_i \in \mathcal{C}} |\mathcal{C}_i|$.
    \item \textit{Medium entropy}: There is some ambiguity across top candidates. We use the frontier LLM as a judge to select from the top-2 clusters.
    \item \textit{High entropy}: High disagreement across generations necessitates either unifying top translations or abstaining from using the sample.
\end{itemize}

\textbf{Knowledge Distillation.}
Following the pipeline above, we obtain a final dataset $\left\{(\vp_i, \vt_i)\right\}_{i=1}^{N}$. We fine-tune a smaller student model (e.g., \texttt{Qwen2.5-7B}) using LoRA, distilling knowledge from the frontier LLM.

%% file: sections/exp.tex
% \newpage
\vspace{-0.1in}
\section{Experiment}
\label{sec:exp}
\vspace{-0.1in}
\subsection{Experiment Settings}

We conduct experiments to evaluate the translation of natural language prompts into SMT-LIB formulas under the NL2SMT framework.

\textbf{Datasets.}
We evaluate our method on 
% two complementary NL2SMT datasets: Customer Service and 
NL2SMT FOLIO dataset. 
% The Customer Service dataset is synthetically generated from policy documents and contains customer-agent conversations paired with their corresponding SMT-LIB translations. 
FOLIO is an open-domain first-order logic reasoning benchmark comprising natural language premises and conclusions paired with SMT-LIB assertions constructed using predefined logical declarations. Together, we evaluate NL2SMT translation in both realistic policy-oriented conversations and controlled logical-reasoning scenarios.

\textbf{Training Strategy.}
We fine-tune pretrained language models using \emph{LoRA} (Low-Rank Adaptation), a parameter-efficient fine-tuning method. We use a learning rate of $5 \times 10^{-5}$, a batch size of 32, a LoRA rank of 32, and a LoRA scaling factor $\alpha$ of 64.

\textbf{Evaluated Models.}
Our primary student model is Qwen2.5-7B-Instruct, on which all fine-tuning experiments and ablation studies are conducted. For comparison, we evaluate the few-shot performance of several open-source and proprietary LLMs, including Qwen2.5-7B-Instruct, Qwen3-4B, Qwen3-8B, Qwen3-14B, Mistral-7B-Instruct, and Claude Sonnet 3.7.

\textbf{Evaluation Metrics.}
We adopt two complementary metrics to evaluate translation quality. First, for the \emph{Binary Equivalence Check}, we use the Z3 theorem prover to determine whether a generated SMT-LIB formula is logically equivalent to the ground-truth formula. We report Pass@10, which measures whether at least one of ten generated candidates passes the equivalence check. Second, for the \emph{Continuous Similarity Score}, we measure partial logical similarity in the range $[0,1]$ when exact equivalence is not achieved. This score is computed through symbolic compression and anti-unification of the generated and ground-truth SMT-LIB formulas using an SMT-LIB specification implemented in Egglog.

\subsection{Logical NL2SMT Translation}

% Keep the numerical contents of the original table unchanged.
\begin{table}[h!]
\centering
\vspace{-0.2in} 
\caption{Pass@K scores (\%) for FOLIO Translation.}
\setlength{\tabcolsep}{2pt}
% \small
\resizebox{\textwidth}{!}{
\begin{tabular}{lcccccccccc}
\toprule
\textbf{Model} & \textbf{Pass@10} & \textbf{Pass@9} & \textbf{Pass@8} & \textbf{Pass@7} & \textbf{Pass@6} & \textbf{Pass@5} & \textbf{Pass@4} & \textbf{Pass@3} & \textbf{Pass@2} & \textbf{Pass@1} \\
\midrule
Qwen3-4B & 19.792 & 19.792 & 19.792 & 19.792 & 19.792 & 19.792 & 19.792 & 19.792 & 18.750 & 17.708 \\
Qwen3-8B & 18.750 & 18.750 & 18.750 & 18.750 & 18.750 & 18.750 & 16.667 & 16.667 & 16.667 & 16.667 \\
Qwen3-14B & 42.708 & 42.708 & 42.708 & 42.708 & 42.708 & 42.708 & 42.708 & 42.708 & 42.708 & 42.708 \\
Mistral-7B-Instruct & 6.250 & 6.250 & 6.250 & 6.250 & 6.250 & 5.208 & 5.208 & 4.167 & 4.167 & 4.167 \\
Qwen2.5-7B-Instruct & 21.875 & 21.875 & 21.875 & 21.875 & 20.833 & 19.792 & 18.750 & 18.750 & 18.750 & 15.625 \\
\midrule
Distillation & 50.347 & 50.347 & 49.306 & 48.264 & 47.222 & 46.181 & 45.486 & 43.750 & 42.708 & 39.931 \\
% SCD (Majority)  & 53.819 & 53.819 & 53.819 & 53.472 & 51.042 & 48.611 & 47.222 & 46.181 & 42.361 & 38.542 \
% SCD (Low-E) & 46.875 & 46.875 & 46.875 & 45.833 & 43.403 & 42.014 & 41.389 & 38.194 & 32.639 & 30.208 \
% SCD (High-E) & 47.569 & 47.569 & 47.222 & 46.875 & 43.403 & 42.708 & 40.972 & 39.236 & 35.417 & 35.417 \
% SCD (Refine-Abstain) & 52.083 & 52.083 & 51.736 & 50.694 & 49.653 & 48.958 & 46.875 & 45.486 & 44.097 & 42.361 \
SCD & 55.208 & 55.208 & 54.514 & 52.778 & 50.083 & 49.653 & 48.958 &
47.917	& 46.528 &	44.097 \\
\bottomrule
\end{tabular}
}

\label{tab:pass_k_folio}
\end{table}

\input{sections/exp_impli}
\textbf{FOLIO Dataset.}
% In addition to the Customer Service dataset, 
We evaluate our method on FOLIO and report Pass@K for $K=1,\ldots,10$ in Table~\ref{tab:pass_k_folio}. We make the following observations:

\begin{itemize}[left=0.0em]
    \item Both distillation methods substantially outperform the pretrained baselines. For example, Qwen2.5-7B-Instruct achieves a Pass@10 of 21.875\% in the few-shot setting, whereas vanilla distillation improves it to 50.347\%.

    \item Vanilla distillation also outperforms the strongest pretrained baseline, Qwen3-14B, by 7.639 percentage points in Pass@10 (50.347\% versus 42.708\%), despite using a smaller student model.

    \item SCD achieves the best Pass@10 performance of 55.208\%, exceeding Qwen3-14B by 12.500 percentage points and vanilla distillation by 4.861 percentage points.
\end{itemize}

% \newpage
\subsection{Downstream Analysis}

\begin{figure}[h!]
    \centering
    \includegraphics[width=0.9\textwidth]{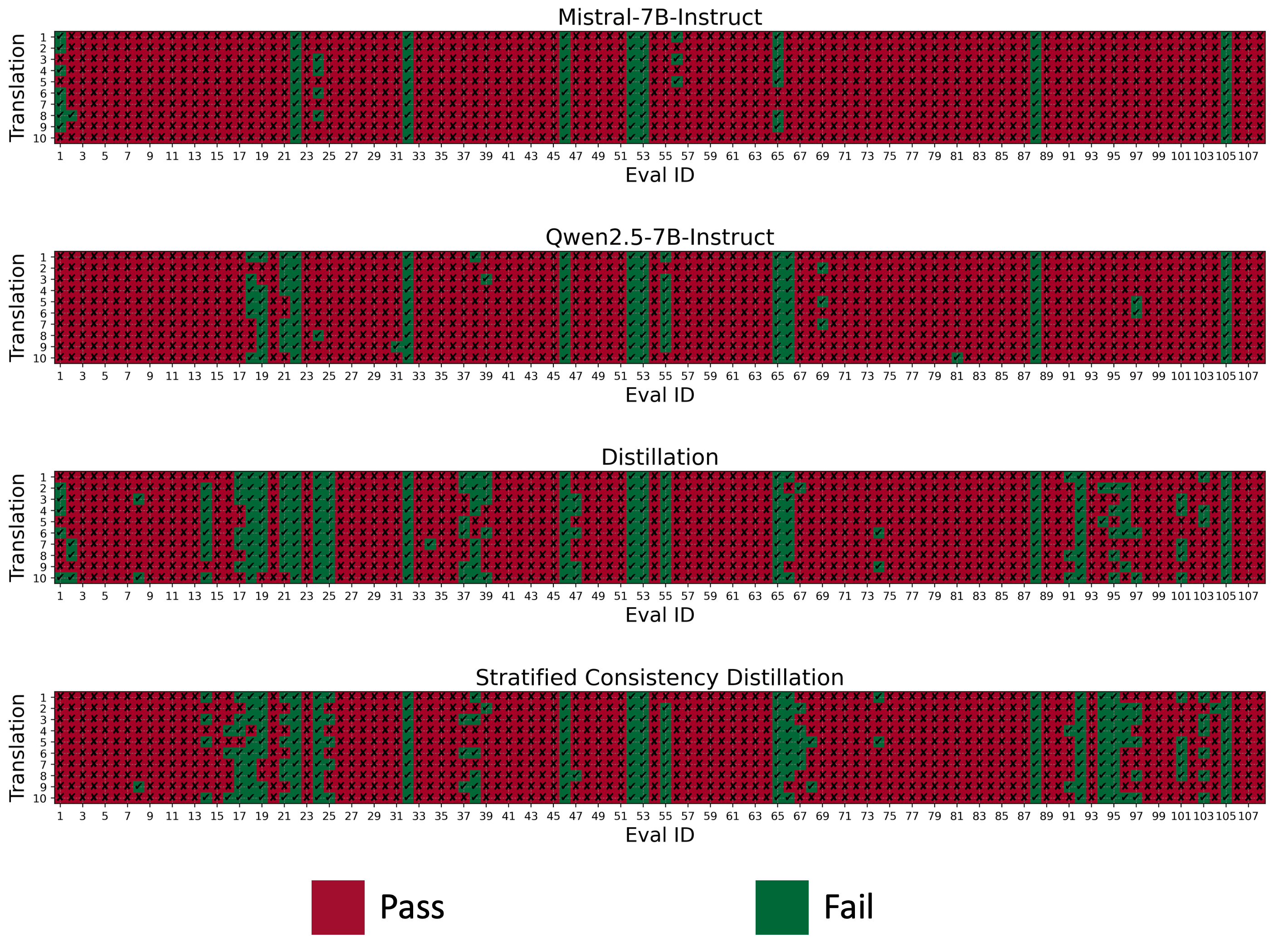}
    \vspace{-0.1in}
    \caption{Per-example Pass@10 results under different model and training configurations.}
    \label{fig:visual_pass10}
    \vspace{-0.1in}
\end{figure}

\begin{figure}[h!]
    \centering
    \includegraphics[width=0.9\textwidth]{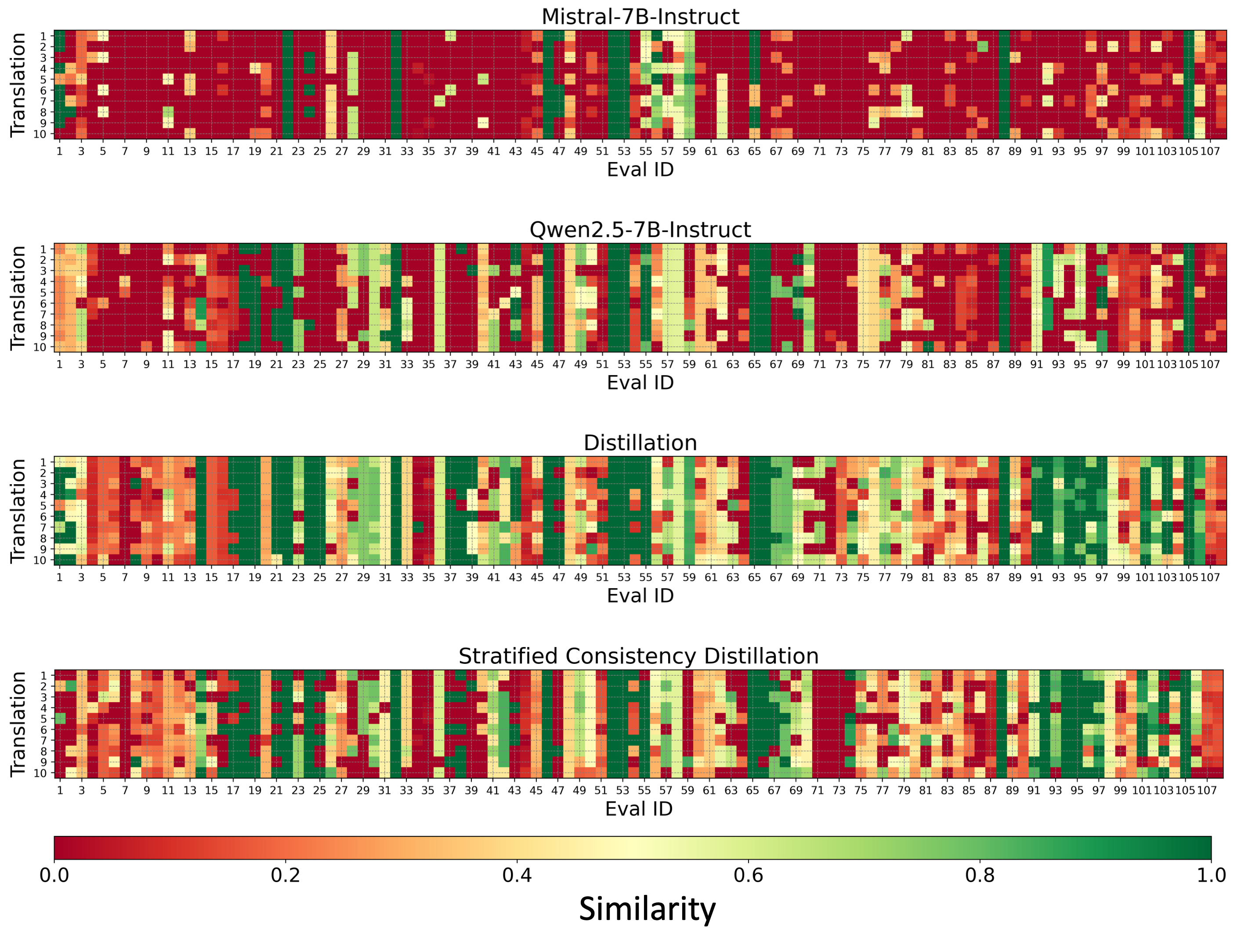}
    \vspace{-0.1in}
    \caption{Per-example similarity scores under different model and training configurations.}
    \vspace{-0.1in}
    \label{fig:visual_sim}
\end{figure}

\textbf{Visualization.}
Figures~\ref{fig:visual_pass10} and~\ref{fig:visual_sim} visualize the per-example Pass@10 outcomes and continuous similarity scores, respectively, under different model and training configurations. In Figure~\ref{fig:visual_pass10}, red denotes an unsuccessful translation and green denotes a successful translation. In Figure~\ref{fig:visual_sim}, colors range from red for low similarity to green for high similarity.
The evaluated configurations include pretrained Mistral-7B-Instruct, pretrained Qwen2.5-7B-Instruct, Qwen2.5-7B-Instruct fine-tuned on NL2SMT dataset, and the corresponding model trained using our consistency-distillation framework. The Pass@10 heatmaps show a gradual transition from predominantly unsuccessful predictions for the pretrained models to broader coverage of correct translations after fine-tuning and distillation. Mistral-7B-Instruct exhibits the sparsest coverage of successful predictions, followed by pretrained Qwen2.5-7B-Instruct. Fine-tuning on the NL2SMT dataset increases the number of correctly translated examples, while consistency distillation produces the broadest coverage.
The similarity heatmaps exhibit a comparable trend. The pretrained models contain larger regions of low similarity, whereas fine-tuning and consistency distillation shift the distribution toward higher similarity values, indicating stronger logical alignment with the ground-truth translations.

% Keep the numerical contents of the original latency table unchanged.
% \begin{table}[h!]
\begin{wraptable}{r}{0.5\textwidth}
\centering
% \small
\vspace{-0.17in} 
\caption{Latency percentiles (P50, P90, P99) of different models on P4d EC2 instance.}
\resizebox{0.5\textwidth}{!}{
\begin{tabular}{lccc}
\toprule
\textbf{Model} & \textbf{P50} & \textbf{P90} & \textbf{P99} \\
\midrule
Claude Sonnet 3.7 & 16.680 & 28.042 & 29.425 \\
Qwen2.5-7B-Instruct & 4.040 & 5.074 & 5.651 \\
Qwen3-4B & 4.294 & 5.698 & 6.281 \\
Qwen3-8B & 2.860 & 4.809 & 4.835 \\
Qwen3-14B & 7.454 & 13.939 & 14.604 \\
Mistral-7B-Instruct-v0.2 & 4.350 & 7.307 & 14.191 \\
\bottomrule
\end{tabular}
}
\vspace{-0.15in} 
\label{tab:latency}
% \end{table}
\end{wraptable}

\textbf{Latency.}
Table~\ref{tab:latency} reports the P50, P90, and P99 inference latency of the evaluated models. The fine-tuned Qwen2.5-7B-Instruct model achieves substantially lower latency than Claude Sonnet 3.7 across all three percentiles. Its P50 latency is 4.040 seconds, approximately $4.1\times$ faster than the 16.680 seconds required by Claude Sonnet 3.7. The same advantage is observed at P90 (5.074 versus 28.042 seconds) and P99 (5.651 versus 29.425 seconds), indicating that the efficiency improvement remains consistent at higher latency percentiles.
Among the open-source models, Qwen2.5-7B-Instruct also provides a favorable efficiency profile. It achieves lower P90 and P99 latency than Qwen3-4B, Qwen3-14B, and Mistral-7B-Instruct, although Qwen3-8B is faster. These results demonstrate that our approach improves translation quality while retaining practical 
% inference 
efficiency.

\begin{figure}[h!]
    \centering
    \vspace{-0.2in} 
    \includegraphics[width=1.0\linewidth]{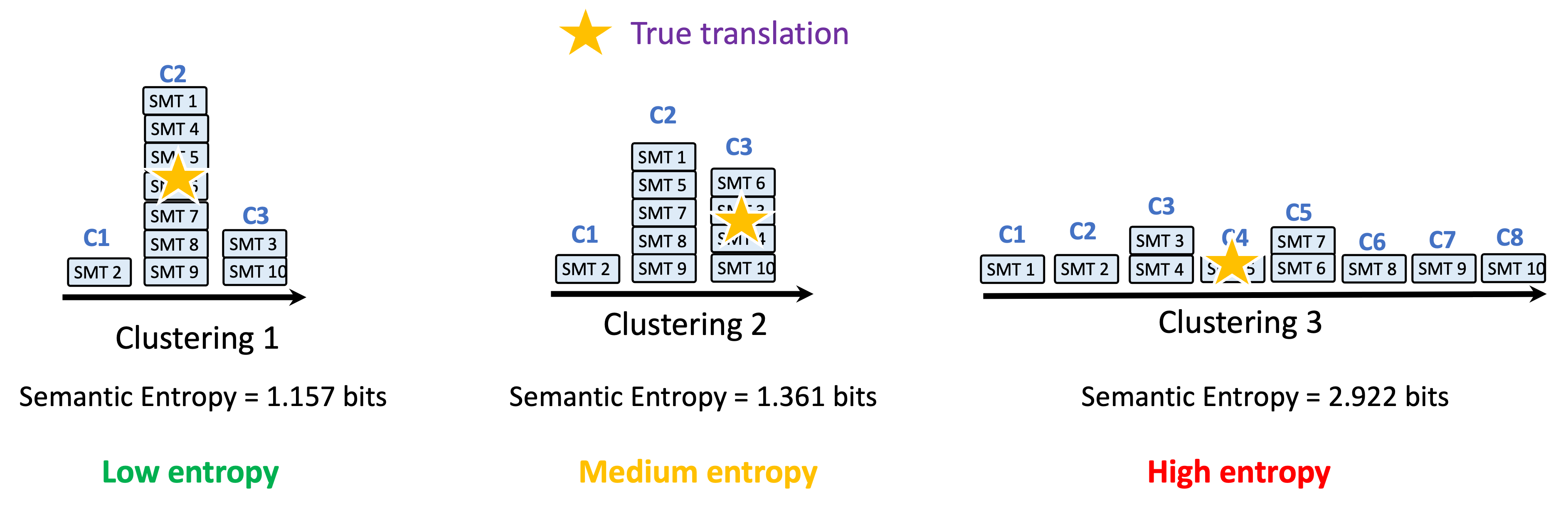}
    \caption{Illustration of semantic-equivalence clusters at different uncertainty levels. Each cluster groups logically equivalent SMT-LIB candidates, and the gold star ($\star$) denotes the ground-truth translation. As entropy increases, the candidates become more evenly distributed across clusters, reducing the reliability of the largest cluster. (a) \textbf{Low entropy}: the largest cluster contains the ground truth; (b) \textbf{Medium entropy}: the ground truth appears in a secondary cluster; and (c) \textbf{High entropy}: the largest cluster is no longer a reliable indicator of correctness.}
    \label{fig:three_entropy}
\end{figure}

\textbf{Reliability of Symbolic Semantic Entropy.}
To validate the motivation for stratified consistency distillation, we investigate whether the sizes of semantic-equivalence clusters provide a reliable signal for selecting the correct translation. We treat \emph{symbolic semantic entropy} as an indicator of prediction uncertainty, as illustrated in Figure~\ref{fig:three_entropy}. In the low-entropy regime, the largest cluster is dominant and is therefore likely to contain the correct translation. In the medium-entropy regime, the correct translation may instead occur in a secondary cluster. In the high-entropy regime, candidate translations are more evenly distributed across clusters, making the largest cluster less reliable.
Table~\ref{tab:unification} compares different cluster-based selection strategies. Selecting the largest cluster (\textbf{Top@1}) performs better than randomly selecting a candidate (\textbf{Pass@1}). Moreover, \textbf{Top@2} approaches the oracle upper bound represented by \textbf{Pass@10}, indicating that the correct translation is usually contained within one of the two largest clusters.

\begin{table}[h!]
    \centering
    \vspace{-0.2in} 
    \caption{Performance of cluster-based candidate-selection. \textbf{Pass@1 (Random)} randomly selects one candidate, whereas \textbf{Top@$k$} considers candidates from the $k$ largest semantic-equivalence clusters.}
\resizebox{0.8\textwidth}{!}{
    \begin{tabular}{lccccc}
        \toprule
        \textbf{Model} & \textbf{Pass@10} & \textbf{Pass@1 (Random)} & \textbf{Top@1} & \textbf{Top@2} & \textbf{Top@3} \\
        \midrule
        Qwen2.5-7B & 17.593 & 13.889 & 15.741 & 17.593 & 17.593 \\
        SCD & 31.481 & 24.074 & 25.926 & 30.556 & 31.481 \\
        \bottomrule
    \end{tabular}
}
    \vspace{-0.15in} 
    \label{tab:unification}
\end{table}

\textbf{Stratified Consistency Distillation.}
The ablation results in Table~\ref{tab:ablation_scd_customer_service} show that the distillation strategies substantially improve Pass@K over the pretrained Qwen2.5-7B baseline. Vanilla distillation already produces a considerable improvement, demonstrating the value of transferring high-quality outputs from the teacher model. Among the entropy-specific variants, training with low-entropy samples consistently outperforms training with high-entropy samples across all values of $K$, suggesting that more consistent teacher predictions provide a stronger supervision signal. Most importantly, stratified SCD achieves the highest or tied-highest performance across all reported Pass@K metrics. This result demonstrates that applying different selection strategies according to semantic entropy produces a more effective and robust distillation signal than relying on a single entropy regime.

\begin{table}[h!]
\centering
\setlength{\tabcolsep}{2pt}
% \small
\vspace{-0.2in} 
\caption{
Ablation study of stratified consistency distillation.}
\resizebox{\textwidth}{!}{
\begin{tabular}{lcccccccccc}
\toprule
\textbf{Model} & \textbf{Pass@10} & \textbf{Pass@9} & \textbf{Pass@8} & \textbf{Pass@7} & \textbf{Pass@6} & \textbf{Pass@5} & \textbf{Pass@4} & \textbf{Pass@3} & \textbf{Pass@2} & \textbf{Pass@1} \\
\midrule
Qwen2.5-7B & 17.593 & 17.593 & 17.593 & 17.593 & 16.667 & 16.667 & 15.741 & 15.741 & 14.815 & 13.889 \\
\midrule
Vanilla Distillation  & 27.469 & 27.469 & 27.116 & 26.543 & 25.926 & 25.617 & 24.383 & 23.148 & 22.222 & 19.753 \\
% Self-Consistency Distillation & 29.012 & 29.012 & 29.012 & 28.086 & 26.852 & 26.852 & 26.235 & 25.309 & 24.074 & 21.914 \
% \midrule
SCD (High Entropy) & 26.235 & 26.235 & 25.000 & 24.691 & 23.765 & 22.679 & 20.679 & 19.753 & 18.519 & 18.519 \\
SCD (Low Entropy) & 28.704 & 28.704 & 28.704 & 28.086 & 27.778 & 26.852 & 25.926 & 25.926 & 24.074 & 23.457 \\
% \midrule
SCD (Stratified) & 31.481 & 31.481 & 31.481 & 30.247 & 29.938 & 26.852 & 26.852 & 26.852 & 26.852 & 26.852 \\
% SCD (Refine-Double) & 28.704 & 28.704 & 28.395 & 27.469 & 26.852 & 26.852 & 25.926 & 24.691 & 22.840 & 20.062 \\
% \midrule
% Only Implication & 29.321 & 29.321 & 29.012 & 28.395 & 27.778 & 27.161 & 26.235 & 24.383 & 20.679 & 20.370 \\
% Auxiliary Learning & 33.024 & 32.716 & 31.790 & 29.938 & 31.716 & 29.938 & 27.778 & 27.716 & 24.383 & 26.235 \\
% Continued Learning (x3) & 37.037 & 37.037 & 36.111 & 35.185 & 34.259 & 33.642 & 32.099 & 30.556 & 29.321 & 26.543 \\
% Continued Learning (x1) & 34.568 & 34.259 & 34.259 & 34.259 & 33.333 & 32.099 & 31.790 & 30.556 & 26.235 & 26.235 \\
% Continued Learning (Active) & 36.420 & 36.420 & 35.802 & 35.802 & 34.876 & 33.950 & 33.642 & 30.864 & 28.395 & 28.395 \\
% Continued Learning (Negative) & 34.259 & 34.259 & 34.259 & 33.333 & 33.024 & 32.062 & 31.481 & 29.630 & 26.235 & 26.235 \\
\bottomrule
\end{tabular}
}

\label{tab:ablation_scd_customer_service}
\end{table}

%% file: sections/related.tex
\vspace{-0.1in} 
\section{Related Works}
\label{sec:related}

\textbf{Improving Logical Reasoning of LLMs.}
Substantial research has been devoted to enhancing the logical reasoning capabilities of large language models (LLMs), with a particular focus on mathematical reasoning tasks. Early studies have shown that pretrained LLMs~\citep{geminiteam2024gemini15unlockingmultimodal,openai2024gpt4technicalreport,anthropic2024claude3} can solve reasoning problems through prompting strategies such as Chain-of-Thought (CoT) reasoning~\citep{fu2022complexity,wei2022chain}, which guides the model to generate intermediate steps before producing the final answer. Beyond prompting, supervised fine-tuning (SFT)~\citep{cobbe2021training, yu2023metamath, lin2026scaling} on high-quality, human-annotated datasets has been shown to yield further improvements in reasoning accuracy. More recently, reinforcement learning from human feedback (RLHF)~\citep{ziegler2019fine, brantley2026accelerating, lin2026scaling, chen2026avoiding} has emerged as a powerful approach for improving reasoning performance, leveraging reward models~\citep{lightman2023let,wang2023math} to align LLM outputs with desired reasoning processes and solutions. However, these techniques for translating informal natural language into formal SMT-LIB formulas remain largely underexplored.

\textbf{Autoformalization of LLMs.}
Autoformalization, the task of converting informal natural language into verifiable formal representations, plays a foundational role in both mathematical formalization and the emerging verification of LLM-generated outputs. In mathematical contexts, it enables the translation of human-written proofs into machine-checkable formats for proof assistants such as Coq~\citep{the_coq_development_team_2024_11551307}, Lean~\citep{de2015lean}, and Isabelle~\citep{ait2008theorem}. More recently, its scope has expanded to address reliability issues in LLMs by translating generated text into precise, logically consistent forms using systems such as first-order logic~\citep{ryu2024divide} or arithmetic frameworks like Peano arithmetic~\citep{kennedy1974giuseppe}. By bridging the expressive flexibility of natural language and the rigor of formal verification, autoformalization mitigates semantic ambiguity and supports robust reasoning. However, translating LLM outputs directly into SMT-LIB—a critical formalism for automated reasoning over logical constraints—remains largely unexplored. This gap motivates our work, which aims to advance LLM automatization in this under-addressed direction.

%% file: sections/conclusion.tex
\section{Conclusion}
\label{sec:conclusion}

% In this work, we addressed the challenge of ensuring the logical soundness and policy compliance of LLM-generated responses in high-stakes domains. We proposed a principled NL2SMT framework that translates natural language into verifiable SMT-LIB formulas, enabling automated verification using symbolic solvers. Our Stratified Consistency Distillation method selectively distills the logical translation capabilities of a frontier teacher model into a smaller open-source student model based on the uncertainty of its generated translations. Experiments on the Customer Service and FOLIO datasets demonstrate that our method consistently improves translation accuracy over pretrained and vanilla-distillation baselines while achieving competitive or superior performance to substantially larger models on key metrics. Moreover, the distilled model provides considerably lower inference latency, making the framework more practical for real-world deployment. Overall, this work provides a scalable approach to developing efficient, reliable, and verifiable language models, with broad applicability to domains in which logical correctness and policy compliance are essential.

In this work, we addressed the challenge of ensuring the logical soundness of
LLM-generated responses in high-stakes domains. We proposed a principled NL2SMT
framework that translates natural language into verifiable SMT-LIB formulas,
enabling automated verification with symbolic solvers. Our Stratified
Consistency Distillation method selectively distills the logical translation
capability of a frontier teacher model into a smaller open-source student,
allocating supervision according to the uncertainty of the generated
translations. Experiments on FOLIO show that our method improves translation
accuracy over both pretrained and vanilla-distillation baselines, and matches or
exceeds substantially larger models despite using far fewer parameters. The
distilled model further offers considerably lower inference latency, making the
framework more practical to deploy. Overall, this work provides a scalable route
to efficient, reliable, and verifiable language models for domains in which
logical correctness is essential.

%% file: sections/app.tex
\newpage
\section{Prompt for Translation Generation}
\label{sec:prompt}

\begin{tcolorbox}[
title=Prompt for Translation Generation (FOLIO),
colback=gray!5!white,
colframe=black!75,
fonttitle=\bfseries,
breakable]

% \centering\textbf{Meta SMT $\varphi$}
\begin{lstlisting}[style=smt]
You are an expert in autoformalization - translating natural language logical reasoning into SMT-LIB format.

Your task: Given natural language premises and conclusion, generate SMT-LIB assertions using the provided declarations.

<instructions>
1. Analyze the SMT-LIB variables provided within the 'declarations' section to understand all allowed variables, their types, and descriptions.
2. Translate the natural language premises and conclusion into SMT-LIB assertions
3. Use only the constants and functions declared in the declarations section
4. Each assertion should be wrapped in (assert ...)
5. Use proper SMT-LIB syntax for logical operators: and, or, not, =>, xor, forall, exists
6. DO NOT include any XML tags or markdown formatting - output only pure SMT-LIB assertions
</instructions>

Here is an example:

<example>
<input>
Premises: All people who regularly drink coffee are dependent on caffeine. People regularly drink coffee, or they don't want to be addicted to caffeine, or both. No one who doesn't want to be addicted to caffeine is aware that caffeine is a drug. Rina is either a student or unaware that caffeine is a drug, but not both. Rina is either dependent on caffeine or a student, but not both.
Conclusion: Rina doesn't want to be addicted to caffeine or is unaware that caffeine is a drug.
</input>

<declarations>
(declare-sort Individual)
(declare-const rina Individual)
(declare-const coffee Individual)
(declare-const caffeine Individual)
(declare-fun DrinkRegularly (Individual Individual) Bool)
(declare-fun IsDependentOn (Individual Individual) Bool)
(declare-fun WantToBeAddictedTo (Individual Individual) Bool)
(declare-fun AwareThatDrug (Individual Individual) Bool)
(declare-fun Student (Individual) Bool)
</declarations>

Output (pure SMT-LIB assertions only):
(assert (forall ((x Individual)) (=> (DrinkRegularly x coffee) (IsDependentOn x caffeine))))
(assert (forall ((x Individual)) (or (DrinkRegularly x coffee) (not (WantToBeAddictedTo x caffeine)))))
(assert (forall ((x Individual)) (=> (not (WantToBeAddictedTo x caffeine)) (not (AwareThatDrug x caffeine)))))
(assert (not (xor (Student rina) (not (AwareThatDrug rina caffeine)))))
(assert (not (xor (IsDependentOn rina caffeine) (Student rina))))
(assert (or (not (WantToBeAddictedTo rina caffeine)) (not (AwareThatDrug rina caffeine))))
</example>

Now translate this natural language input using the provided declarations:

<input>
Premises: No sandwich cookies are healthy.
Oreos are sandwich cookies.
Conclusion: All sandwich cookies are delicious.
</input>

<declarations>
(declare-sort Individual)
(declare-const oreos Individual)
(declare-fun SandwichCookie (Individual) Bool)
(declare-fun Healthy (Individual) Bool)
(declare-fun Delicious (Individual) Bool)
</declarations>

\end{lstlisting}

\end{tcolorbox}

%% file: example_paper.bib
@misc{openai2024gpt4technicalreport,
  title     = {GPT-4 Technical Report},
  author    = {OpenAI and Josh Achiam et al.},
  year      = {2024},
  eprint    = {2303.08774},
  archivePrefix = {arXiv},
  primaryClass  = {cs.CL},
  url       = {https://arxiv.org/abs/2303.08774}
}

@article{farquhar2024detecting,
  title={Detecting hallucinations in large language models using semantic entropy},
  author={Farquhar, Sebastian and Kossen, Jannik and Kuhn, Lorenz and Gal, Yarin},
  journal={Nature},
  volume={630},
  number={8017},
  pages={625--630},
  year={2024},
  publisher={Nature Publishing Group UK London}
}

@article{ryu2024divide,
  title={Divide and translate: Compositional first-order logic translation and verification for complex logical reasoning},
  author={Ryu, Hyun and Kim, Gyeongman and Lee, Hyemin S and Yang, Eunho},
  journal={arXiv preprint arXiv:2410.08047},
  year={2024}
}

@misc{geminiteam2024gemini15unlockingmultimodal,
      title={Gemini 1.5: Unlocking multimodal understanding across millions of tokens of context}, 
      author={Gemini Team and Petko Georgiev et al.},
      year={2024},
      eprint={2403.05530},
      archivePrefix={arXiv},
      primaryClass={cs.CL},
      url={https://arxiv.org/abs/2403.05530}, 
}

@book{kennedy1974giuseppe,
  title={Giuseppe Peano},
  author={Kennedy, Hubert C and Amsler, Ruth},
  year={1974},
  publisher={Birkh{\"a}user}
}

@inproceedings{de2015lean,
  title={The Lean theorem prover (system description)},
  author={De Moura, Leonardo and Kong, Soonho and Avigad, Jeremy and Van Doorn, Floris and von Raumer, Jakob},
  booktitle={International Conference on Automated Deduction},
  pages={378--388},
  year={2015},
  organization={Springer}
}

@book{ait2008theorem,
  title={Theorem Proving in Higher Order Logics: 21st International Conference, TPHOLs 2008, Montreal, Canada, August 18-21, 2008, Proceedings},
  author={Ait Mohamed, Otmane and Munoz, C{\'e}sar and Tahar, Sofi{\`e}ne},
  volume={5170},
  year={2008},
  publisher={Springer Science \& Business Media}
}

@software{the_coq_development_team_2024_11551307,
  author       = {The Coq Development Team},
  title        = {The Coq Proof Assistant},
  month        = jun,
  year         = 2024,
  publisher    = {Zenodo},
  version      = {8.19},
  doi          = {10.5281/zenodo.11551307},
  url          = {https://doi.org/10.5281/zenodo.11551307},
}

@inproceedings{lightman2023let,
  title={Let's verify step by step},
  author={Lightman, Hunter and Kosaraju, Vineet and Burda, Yuri and Edwards, Harrison and Baker, Bowen and Lee, Teddy and Leike, Jan and Schulman, John and Sutskever, Ilya and Cobbe, Karl},
  booktitle={The Twelfth International Conference on Learning Representations},
  year={2023}
}

@article{wang2023math,
  title={Math-shepherd: Verify and reinforce llms step-by-step without human annotations},
  author={Wang, Peiyi and Li, Lei and Shao, Zhihong and Xu, RX and Dai, Damai and Li, Yifei and Chen, Deli and Wu, Yu and Sui, Zhifang},
  journal={arXiv preprint arXiv:2312.08935},
  year={2023}
}

@article{ziegler2019fine,
  title={Fine-tuning language models from human preferences},
  author={Ziegler, Daniel M and Stiennon, Nisan and Wu, Jeffrey and Brown, Tom B and Radford, Alec and Amodei, Dario and Christiano, Paul and Irving, Geoffrey},
  journal={arXiv preprint arXiv:1909.08593},
  year={2019}
}

@article{cobbe2021training,
  title={Training verifiers to solve math word problems},
  author={Cobbe, Karl and Kosaraju, Vineet and Bavarian, Mohammad and Chen, Mark and Jun, Heewoo and Kaiser, Lukasz and Plappert, Matthias and Tworek, Jerry and Hilton, Jacob and Nakano, Reiichiro and others},
  journal={arXiv preprint arXiv:2110.14168},
  year={2021}
}

@article{yu2023metamath,
  title={Metamath: Bootstrap your own mathematical questions for large language models},
  author={Yu, Longhui and Jiang, Weisen and Shi, Han and Yu, Jincheng and Liu, Zhengying and Zhang, Yu and Kwok, James T and Li, Zhenguo and Weller, Adrian and Liu, Weiyang},
  journal={arXiv preprint arXiv:2309.12284},
  year={2023}
}

@article{fu2022complexity,
  title={Complexity-based prompting for multi-step reasoning},
  author={Fu, Yao and Peng, Hao and Sabharwal, Ashish and Clark, Peter and Khot, Tushar},
  journal={arXiv preprint arXiv:2210.00720},
  year={2022}
}

@article{wei2022chain,
  title={Chain-of-thought prompting elicits reasoning in large language models},
  author={Wei, Jason and Wang, Xuezhi and Schuurmans, Dale and Bosma, Maarten and Xia, Fei and Chi, Ed and Le, Quoc V and Zhou, Denny and others},
  journal={Advances in neural information processing systems},
  volume={35},
  pages={24824--24837},
  year={2022}
}

@misc{anthropic2024claude3,
  author = {Anthropic},
  title = {Claude 3 Model Family},
  year = {2024},
  url = {https://www.anthropic.com/news/claude-3-family},
  note = {Accessed: 2024-10-XX}
}

@misc{fu2023complexitybasedpromptingmultistepreasoning,
      title={Complexity-Based Prompting for Multi-Step Reasoning}, 
      author={Yao Fu and Hao Peng and Ashish Sabharwal and Peter Clark and Tushar Khot},
      year={2023},
      eprint={2210.00720},
      archivePrefix={arXiv},
      primaryClass={cs.CL},
      url={https://arxiv.org/abs/2210.00720}, 
}

@misc{wei2023chainofthoughtpromptingelicitsreasoning,
      title={Chain-of-Thought Prompting Elicits Reasoning in Large Language Models}, 
      author={Jason Wei and Xuezhi Wang and Dale Schuurmans and Maarten Bosma and Brian Ichter and Fei Xia and Ed Chi and Quoc Le and Denny Zhou},
      year={2023},
      eprint={2201.11903},
      archivePrefix={arXiv},
      primaryClass={cs.CL},
      url={https://arxiv.org/abs/2201.11903}, 
}

@article{brantley2026accelerating,
  title={Accelerating rl for llm reasoning with optimal advantage regression},
  author={Brantley, Kiant{\'e} and Chen, Mingyu and Gao, Zhaolin and Lee, Jason and Sun, Wen and Zhan, Wenhao and Zhang, Xuezhou},
  journal={Advances in Neural Information Processing Systems},
  volume={38},
  pages={151492--151531},
  year={2026}
}

@article{lin2026scaling,
  title={Scaling In-Context Online Learning Capability of LLMs via Cross-Episode Meta-RL},
  author={Lin, Xiaofeng and Zhu, Sirou and Chen, Yilei and Chen, Mingyu and Sang, Hejian and Paschalidis, Ioannis and Wang, Zhipeng and Pacchiano, Aldo and Zhang, Xuezhou},
  journal={arXiv preprint arXiv:2602.04089},
  year={2026}
}

@article{chen2026avoiding,
  title={Avoiding exp (R) scaling in RLHF through Preference-based Exploration},
  author={Chen, Mingyu and Chen, Yiding and Sun, Wen and Zhang, Xuezhou},
  journal={Advances in Neural Information Processing Systems},
  volume={38},
  pages={164474--164510},
  year={2026}
}

@article{hou2024protransformer,
  title={Protransformer: Robustify transformers via plug-and-play paradigm},
  author={Hou, Zhichao and Gao, Weizhi and Shen, Yuchen and Wang, Feiyi and Liu, Xiaorui},
  journal={Advances in Neural Information Processing Systems},
  volume={37},
  pages={137557--137609},
  year={2024}
}

@inproceedings{hou2025robustness,
  title={Robustness reprogramming for representation learning},
  author={Hou, Zhichao and Torkamani, MohamadAli and Krim, Hamid and Liu, Xiaorui},
  booktitle={International Conference on Learning Representations},
  volume={2025},
  pages={53682--53704},
  year={2025}
}

@article{gao2025modulated,
  title={Modulated diffusion: Accelerating generative modeling with modulated quantization},
  author={Gao, Weizhi and Hou, Zhichao and Yin, Junqi and Wang, Feiyi and Peng, Linyu and Liu, Xiaorui},
  journal={arXiv preprint arXiv:2506.22463},
  year={2025}
}

@article{hou2026tefm,
  title={TEFM: Token-Efficient Faithful Modeling for Structured Data},
  author={Hou, Zhichao and Sha, Lingdao and Mao, Xueyu and Liu, Yang and Qiu, Peijie and Song, Rui},
  journal={arXiv preprint arXiv:2609.09552},
  year={2026}
}

@article{hou2026towards,
  title={Towards One-for-All Robustness Across a Continuum of Threat Levels},
  author={Hou, Zhichao and Liu, Xiaorui},
  journal={arXiv preprint arXiv:2609.02440},
  year={2026}
}

@article{zhang2026centile,
  title={CENTILE: A Telemetry Foundation Model Evaluated by the Decisions It Drives},
  author={Zhang, Zifan and Hou, Zhichao and Ji, Tingxiang and Liu, Yuchen},
  journal={arXiv preprint arXiv:2608.01725},
  year={2026}
}

@inproceedings{wang2025exploratory,
  title={Exploratory Analysis of the Regulation of Long Non-Coding RNA Transcription with Nucleotide Large Language Models},
  author={Wang, Wei and Hou, Zhichao and Wu, Tianfu and Liu, Xiaorui and Peng, Xinxia},
  booktitle={2025 IEEE International Conference on Bioinformatics and Biomedicine (BIBM)},
  pages={7764--7771},
  year={2025},
  organization={IEEE}
}
